\documentclass[letterpaper, 10 pt, conference]{ieeeconf} 
\IEEEoverridecommandlockouts
\usepackage{times}
\usepackage[T1]{fontenc}
\usepackage{exscale}
\usepackage[centertags]{amsmath}
\usepackage{graphicx}
\usepackage{color}
\usepackage{amsmath}
\usepackage{amssymb}
\usepackage{rotating}
\usepackage{multirow}
\usepackage{subfigure}
\usepackage{caption}

\newcommand{\bitem}{\begin{itemize}}
\newcommand{\eitem}{\end{itemize}}
\newcommand{\benum}{\begin{enumerate}}
\newcommand{\eenum}{\end{enumerate}}

\AtBeginDocument{%

}

\newcommand{\amarsi}[1]{AMARSi}
\newcommand{\dsl}[1]{domain-specific language}
\newcommand{\DSL}[1]{Domain-Specific Language}

\newcommand{\graphdsl}[1]{Graph DSL}
\newcommand{\ccadsl}[1]{Component DSL}
\newcommand{\amdsl}[1]{\amarsi{} DSL} 

\graphicspath{{figs/}}

\title{\LARGE \bf A
\DSL{} for Rich Motor Skill Architectures
}

\author{Arne Nordmann and Sebastian Wrede
\thanks{A. Nordmann and S. Wrede are with the Research
Institute for Cognition and Robotics, Bielefeld University, P.O. Box 100131,
Bielefeld, Germany
{\tt\small \{anordman,swrede\}@cor-lab.uni-bielefeld.de}}%
}

\begin{document}

\maketitle
\thispagestyle{empty}
\pagestyle{empty}

\begin{abstract}
Model-driven software development is a promising way to cope with the complexity
of system integration in advanced robotics, as it already demonstrated its
benefits in domains with comparably challenging system integration requirements.
This paper reports on work in progress in this area which aims to improve the
research and experimentation process in a collaborative research project
developing motor skill architectures for compliant robots. Our goal is to establish a model-driven
development process throughout the project around a \dsl{} (DSL) facilitating the
compact description of adaptive modular architectures for rich motor skills.
Incorporating further languages for other aspects (e.g. mapping to a technical
component architecture) the approach allows not only the formal description of
motor skill architectures but also automated code-generation for
experimentation on technical robot platforms. This paper reports on a first
case study exemplifying how the developed \amdsl{} helps to conceptualize
different architectural approaches and to identify their similarities and differences.
\end{abstract}

\section{\uppercase{Introduction}}
\label{sec:introduction}
\begin{figure}[t]
  \includegraphics[width=0.21\textwidth]{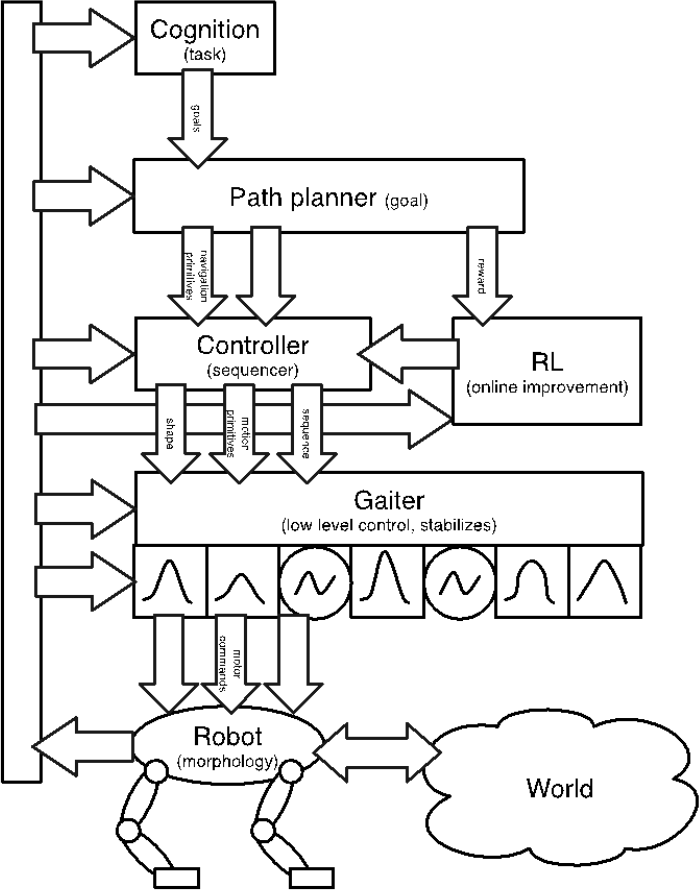}
  \hfill
  \includegraphics[width=0.20\textwidth]{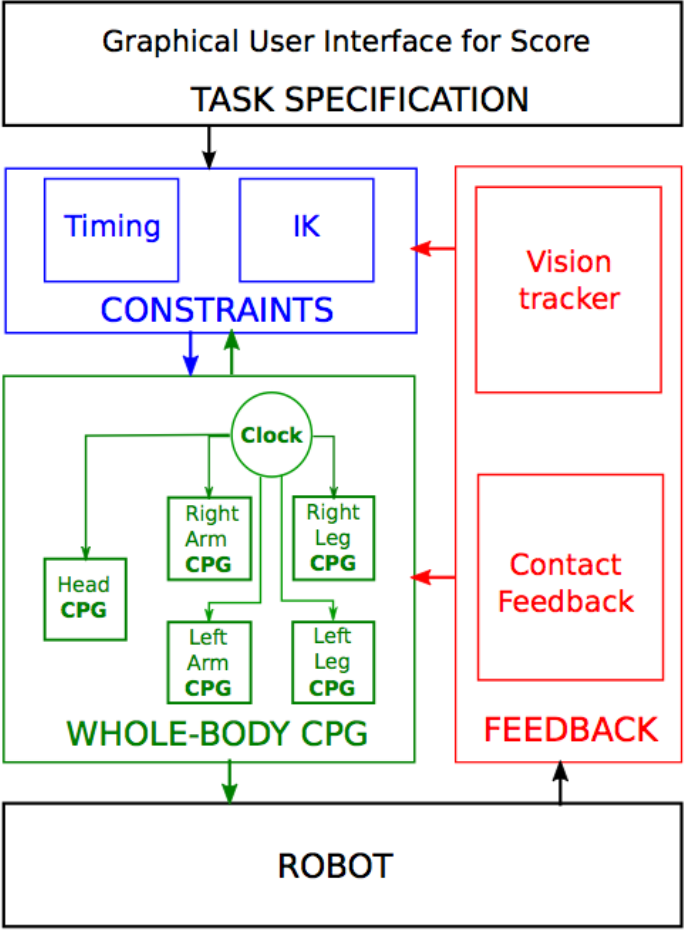}
  \\[0.5em]
  \includegraphics[width=0.30\textwidth]{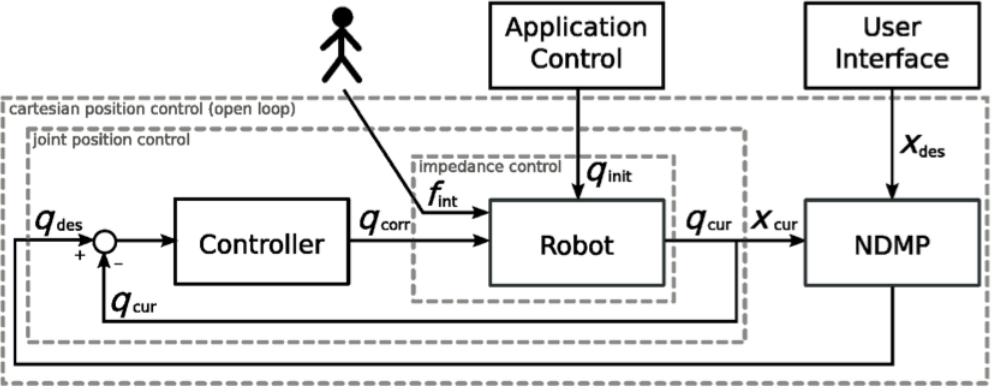}
  \hfill
  \includegraphics[width=0.16\textwidth]{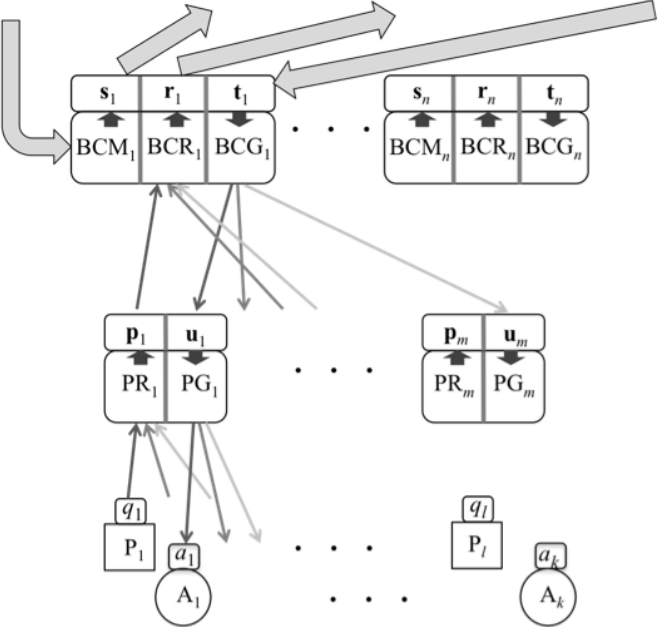}
  \caption{Different motor control architectures from the \amarsi{} project
 expressed in different notations, which makes them hard to compare.}
 \label{fig:example:diff}
\end{figure}

The European \emph{\amarsi{}} Project\footnote{http://www.amarsi-project.eu}
explores how to extend robot motor skills towards biological richness through
the development of modular motor control architectures that thoroughly integrate
compliance, learning and adaptive control. A number of inter-disciplinary
project partners from Biology,
Machine Learning, Systems Engineering and more conduct experimental research to
developed the required architectural concepts based on the idea of modeling
movement primitives as dynamical systems.
Despite this shared approach, a large number of individual architectural designs
on how to organize movement primitives existed at the start of the project.
Although they all describe systems of the same domain, they are expressed in
different notations and blueprints, thus making them hard to compare, see
Figure~\ref{fig:example:diff}.

The discussion about the semantic interpretation of the individual boxologies
raised the questions on how to i) enable research on motor control
architectures among project partners in a way that shares the same conceptual
models and ii) how to link these to reproducable robotics experiments on the
compliant robot platforms. Throughout this paper and as the initial use-case for the
developed DSL on rich motor skills, we focus on the identification of the
domain-specific concepts and its application to improve the interaction among
project participants.

Model-driven and domain specific development methods are recognized to cope
with the challenges of building complex heterogeneous systems in domains such
as aerospace, telecommunication and automotive~\cite{VanDeursen2000} which face
similarly complex integration and modeling challenges as advanced robotics.
In the last years, this approach was initially adapted to the robotics domain,
for example for task description~\cite{Pembeci2002,Simmons1998}.
These approaches are often based on rather generic domain models or
ontologies~\cite{Lortal2011} for semantic modeling of robotics systems and
translated into DSLs~\cite{Zhang2012}. 
Our work in the \amarsi{} project targets at establishing a model-driven
development process focussed on motor control architectures based on movement
primitives. To the best of our knowledge no such process or domain-specific
language for motor control in advanced robots exists so far. 
The initial conceptualization achieved in this work mainly considers structural
aspects but already helped to understand and discuss the semantics of different
architectural elements used by project partners.

The paper is organized as follows:
Section~\ref{sec:da} discusses basic requirements for an DSL in the \amarsi{}
motor control domain, based on a domain analysis and a domain-specific
meta-model.
Based on this analysis, Section~\ref{sec:dsl} introduces the modularization of
DSLs, focusing on the highest level \amdsl{}, and describing example artifact
generation.
Section~\ref{sec:application} shows an use-case of the presented DSL in the
\amarsi{} project before Section~\ref{sec:conclusion} concludes and provides
an outlook on challenges and further work.

\section{\uppercase{Approach}}
\label{sec:da}
Developing \dsl{}s is usually based on a domain analysis to identify
concepts and common structures, problems and solutions of an application
domain.
Large parts of the subsequently presented DSLs are based on an extensive,
formal domain analysis we conducted in early phases of the project on the
domain of motor skill architectures for compliant robots~\cite{D71}.
The domain we consider here is architectures to generating rich and complex
movements through adaptation and learning, both for goal-directed
(eg. reaching, foot placement) and periodic movements (eg. walking, locomotion).

As a first step we conducted a formal analysis of the domain of compliant robots
control, which led to a detailed view on the control and sensing aspects as
well as domain data types of compliant robot control~\cite{D71}.
The domain analysis was based on the formal \emph{Feature-Oriented
Domain Analysis}~\cite{Cohen1990} method, resulting in, among other results, detailed
feature models for the domain entities.
%
%

A second domain analysis was conducted regarding the different architectural
approaches within the project (cf. Figure~\ref{fig:example:diff}). The results
of these analysis led to the specification of an architectural meta-model.
A meta-model describes the structure and the identified element types of an
application domain. An architectural meta-model adds to this perspective
important concepts from software architecture which are required to structure,
technically orchestrate and deploy complex systems whose overall complexity is
well beyond the comprehension of a single developer.
The meta-model serves as basis for the \dsl{} development, defining rather
generic concepts (e.g. \texttt{Components}, \texttt{Ports}) and
specific ones (e.g. \texttt{Adaptive Component},
\texttt{Tracking Controller})~\cite{D73}.
The \amdsl{} detailed below is based on this meta-model to provide a
specific language and syntax dedicated to the \amarsi{} motor control domain
and their solutions, as well as a type-system based on the
domain analysis on compliant robot control.

\section{\uppercase{AMARSi Domain-Specific Languages}}
\label{sec:dsl}
Current state of our DSL development divided the meta-model
into a set of \dsl{}s, following a \emph{language modularization, extension
and composition} approach~\cite{Voelter2011} (LME\&C), to separate the DSLs by concerns.
The formalization of the meta-model in a \dsl{} will provide an avenue for
automation, e.g., for code generation or automatic model verification.
The introduced languages cover different -- rather technical as well as
semantic -- aspects of the domain.

\begin{figure}[t]
 \centering
 \includegraphics[width=.35\textwidth]{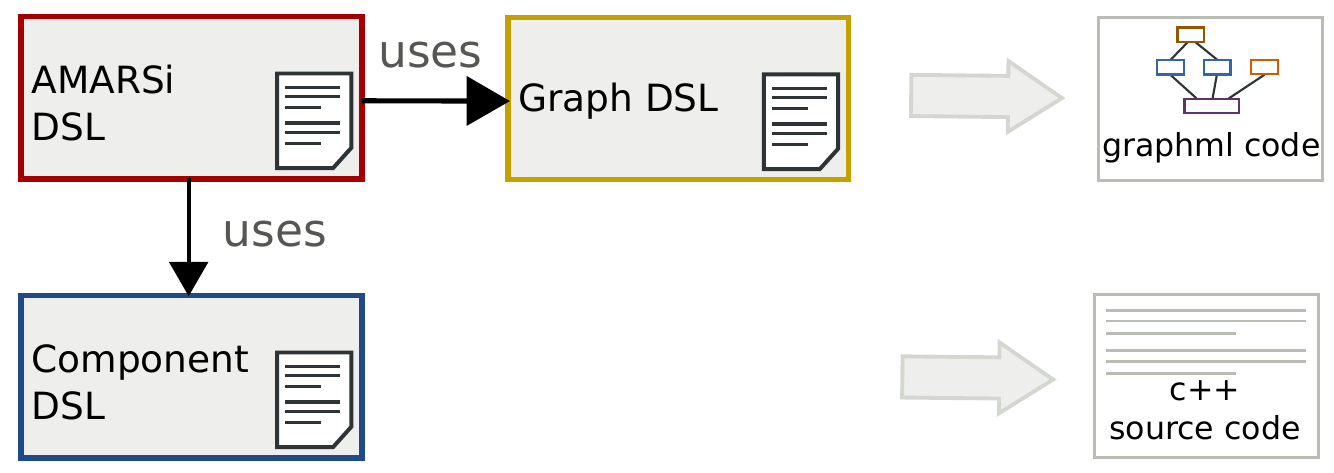}
 \caption{Language decomposition and artifacts generation from the highest
 level \amdsl{} language by using the \ccadsl{} and \graphdsl{}.}
 \label{fig:lmce}
\end{figure}

The languages were developed together with their projectional editors, using the
DSL workbench \emph{Meta-Programming System}~\cite{Voelter2010} (MPS) by
\emph{JetBrains}\footnote{http://www.jetbrains.com/mps/}.
The projectional editors allow to write systems in concepts,
which are \emph{projected} to text for editing purposes~\cite{Voelter2010}.
This saves additional parsing steps of the language to extract the concepts
from text.

\subsection{Language Modularization}
\label{sec:dsl:lcme}

Decomposing the meta-model into a set of \dsl{}s led to the following DSL prototypes:

\benum{}
\item \textbf{\amdsl{}}
 Expresses motor control systems in the concepts found in the domain analysis,
 organized around the essential concept of \amarsi{} \texttt{Adaptive
 Module}. It focuses solely on the domain concepts, hiding the technical
 aspects.
 Concepts of this language cover the architectural motor control aspects as
 well as a type-system and are detailed in the following Section~\ref{sec:dsl:am}.
\item \textbf{\ccadsl{}}
 Covering the aspect of organizing the system in \emph{components} and their
 connectivity.
 Concepts of this language cover component-based software aspects like
 \texttt{Components} with \texttt{States}, \texttt{Ports}, etc.
\item \textbf{\graphdsl{}}
 To cover the aspect of \emph{visualization} of complex robotics architectures
 we follow the already established \emph{GraphML} syntax to express systems
 in the concepts of \texttt{Nodes}, \texttt{Edges} and \texttt{Subgraphs}, as
 well as optical aspects such as shapes, colors, etc.
 The following system visualizations in Figure~\ref{fig:example:dsl} are
 rendered based on the generators of this language.
\eenum{}

\subsection{\amdsl{}}
\label{sec:dsl:am}

The \amdsl{} is created to
allow compact and rich description of motor control architectures in terms of
movement primitives using dynamical systems.
The \amdsl{} is created around the essential concepts of the domain,
\texttt{Adaptive Modules}, \texttt{Adaptive Components}, \texttt{Spaces} and
\texttt{Mappings}, with \texttt{Adaptive Modules} being the architectural
building block to represent a movement primitive.
Movement primitives in \amarsi{} are based on \texttt{Dynamical
Systems}, potentially adapted and trained with machine learning methods.
The essential concepts of the \amdsl{} are:

\begin{figure*}[t]
 \centering
 \includegraphics[width=.95\textwidth]{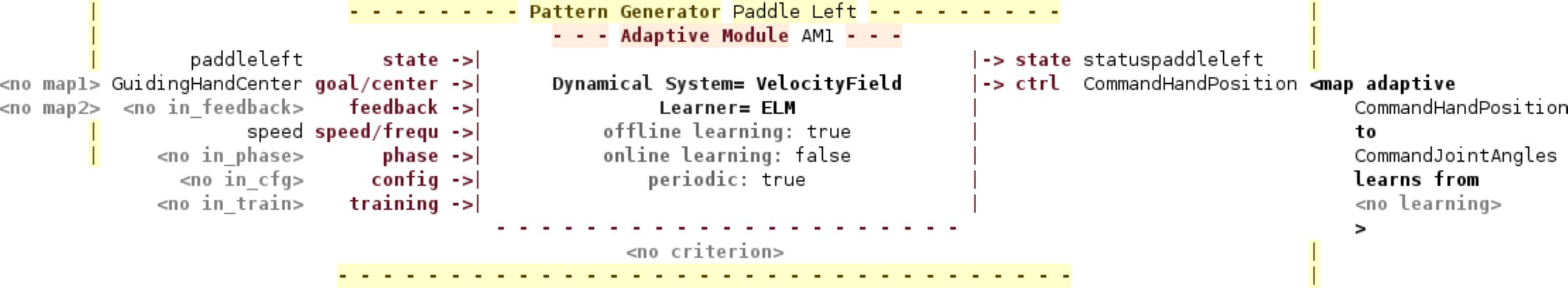}
 \caption{Example of an Adaptive Component in teh projectional \amdsl{} editor.}
 \label{fig:sandbox:am}
\end{figure*}

\bitem{}

\item A \texttt{Space} is as \emph{``a number of explicit variables that appear to be
jointly manipulated or sensed somewhere in our motion control architecture''},
e.g. joint angles of a certain robot limb~\cite{D73}. The \texttt{Types} of
\texttt{Spaces} are the ones found in the robot control domain analysis introduced
before (e.g. \texttt{Joint Angles}, \texttt{Impedance}).

\item An \texttt{Adaptive Module} is the main functional building block of the
motor control system, representing a movement primitive, containing one or
more \texttt{Dynamical Systems} and providing an appropriate \texttt{Learner}
(learning method) for them. It can be parametrized in shape, speed and goal,
its learning methodology (e.g. online, offline) through according input \texttt{Spaces}, and can operate closed-loop or
open-loop. It defines input \texttt{Spaces} for execution and for learning as
well as an output space for its control output.

\item An \texttt{Adaptive Component} is an \emph{``adaptive module together with its
input and output Spaces, a basic semantics (control logic) inside the component
and timing management"}~\cite{D73}. It provides the logical structure around an
\texttt{Adaptive Module}, connecting it with the entire system and managing
connections through different states (online learning, offline learning,
execution).
It can also optionally specify a \texttt{Criterion} that indicates, when the
movement primitive is finished.
The DSL defines several specialized subtypes of \texttt{Adaptive Components}
found during the domain analysis, like \texttt{Tracking Controller},
\texttt{Sequencer} or \texttt{Pattern Generator}, with specific logical wiring.

\item \texttt{Mappings} map data between \texttt{Spaces} of different
\texttt{Type} (e.g. \texttt{Forward Kinematics}), \texttt{Transformations}
define the transformation of data between \texttt{Spaces} of the same
\texttt{Type} (e.g. \texttt{Coordinate Transformation}).

\eitem{}

An example of a movement primitive to perform a paddling movement with the left
arm of a humanoid robot is shown in Figure~\ref{fig:sandbox:am}.
The \texttt{Adaptive Module} (red) contains a \texttt{Velocity Field} as
\texttt{Dynamical System}, that is trained with an
\texttt{Extreme Learning Machine} as \texttt{Learner}. The movement can be
shaped or adapted for example in its \texttt{Speed} or \texttt{Goal} as
indicated by the input \texttt{Spaces} on the left hand side.
The \texttt{Adaptive Module} is embedded in an \texttt{Adaptive Component} (yellow) of
the type \texttt{Pattern Generator}, which determines the internal wiring of
the \texttt{Adaptive Module}. It allows to optionally add \texttt{Mappings}
(in the example at the output) or a stop \texttt{Criterion}. The wiring is not
visible in the editor, but determines the mapping to the \ccadsl{} and therefore
the technical component architecture.

An \amarsi{} motor control system is modeled by defining \texttt{Spaces},
\texttt{Mappings} and \texttt{Transformations} between them, and connecting
\texttt{Adaptive Modules} and \texttt{Adaptive Components} to the defined
\texttt{Spaces}.
The current state of the projectional editor of the \amdsl{} allows
modeling a system by instantiating and configuring a set of the concepts above.

\subsection{Artifact Generation}
\label{sec:dsl:generation}

Besides modeling systems within the introduced \amdsl{}, we aim on
generating artifacts like visualization and source code from the system model.
Figure~\ref{fig:lmce} shows how the introduced DSLs interplay with each other.
For system visualization, the \amdsl{} uses
the \graphdsl{} by providing a generator to map \amdsl{} concepts to the
\graphdsl{} concepts.
The result is a formulation of a system in the concepts of \texttt{Nodes} and
\texttt{Edges}, enriched by optical aspects like coloring. The \graphdsl{}
itself has a generator rendering \emph{graphml} code for visualization
purposes.

For generation of executable C++ artifacts, the \amdsl{}
language uses the \ccadsl{} to map the system model to software
concepts defined in the \ccadsl{}, like \texttt{Components},
\texttt{Ports}, etc. to express the component connectivity.
Its generator generates C++ code targeting the Compliant Control Architecture
(CCA) and Robot Control Interface (RCI) software libraries~\cite{Nordmann2012}.
It generates executable system files, instantiating and configuring the
components and their ports according to the system specified in the \amdsl{}.
Additionally CCA component hulls are generated for implementation, adding typed
ports and the component life-cycle including (optionally) online and offline learning hooks.

Rather technical aspects, like the deployment configuration, were explicitly
excluded from the \amdsl{} in the LME\&C process, to expose only the motor
control aspects to the \amdsl{} user (the architect).
However, MPS allows to not only generate the transitive artifacts
(e.g. GraphML code, executable C++ code), but also allows to preserve all
intermediate models of the generation process, in our case the \ccadsl{} and
\graphdsl{} models. These intermediate models are generated with either
default values or even missing values in the technical aspects, that are
under-specified in the \amdsl{}.
Missing and default values can then be added and refined by experts of the
particular lower-level DSLs, allowing for example the system administrator to
configure the deployment of the motor control system in the \ccadsl{}, or
allowing to change visual aspects in the \graphdsl{} (eg. coloring, highlighting).

\section{\uppercase{Application}}
\label{sec:application}
As already stated in the motivation of this work, motor control architectures
are expressed at different levels of abstraction, using different notations to
express same concepts, and vice versa. This makes comparing different blueprints
especially hard, which is important to enable research on architectural
level~\cite{D73}.
Even within a single project we identified at least four different motor
control architecture notations, as it was introduced in the beginning and
illustrated in Figure~\ref{fig:example:diff}.

As a first application of the DSL approach and the DSL prototypes introduced
before, we modeled the different architecture blueprints of the project in the
\amdsl{}.
The four motor control architectures introduced in
Figure~\ref{fig:example:diff} for example are now comparable and similarities 
of the approaches, like usage of similar concepts and patterns, and their
difference in structure can be identified easier.

\begin{figure}[t]
  \includegraphics[width=0.25\textwidth]{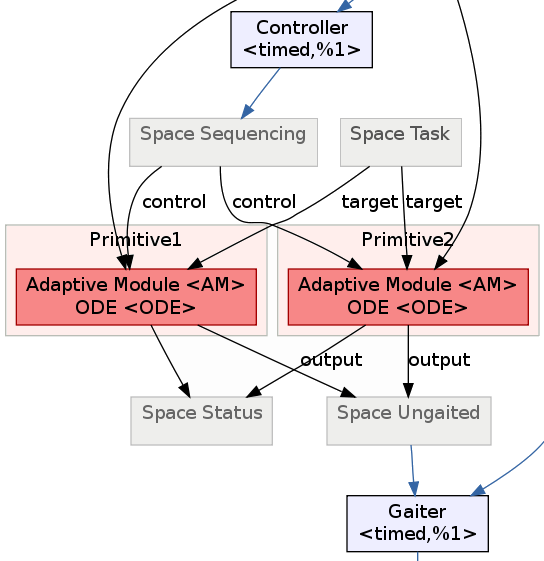}
  \hfill
  \includegraphics[width=0.19\textwidth]{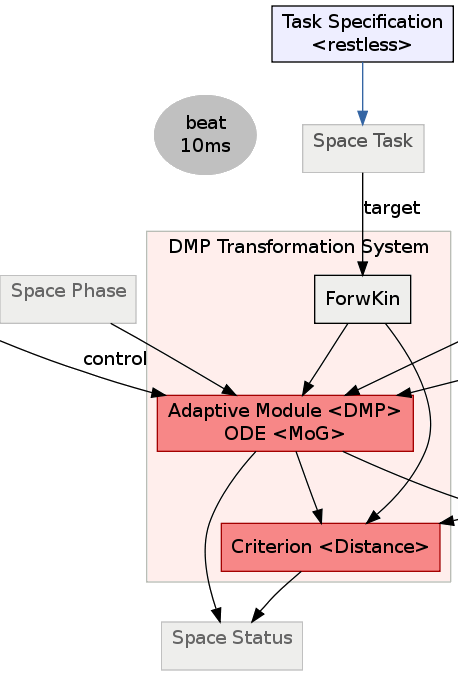}
  \\[0.5em]
  \includegraphics[width=0.28\textwidth]{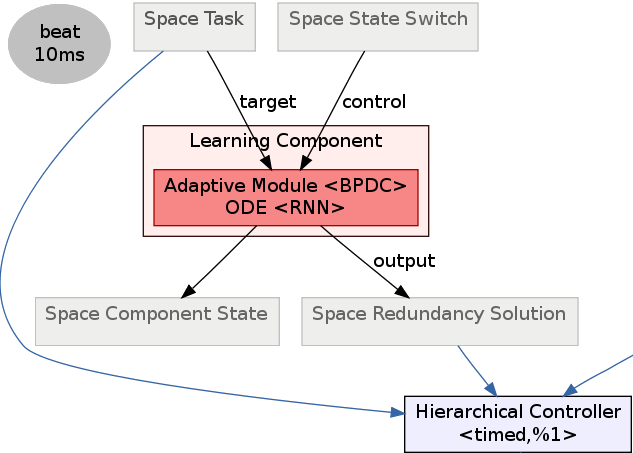}
  \hfill
  \includegraphics[width=0.18\textwidth]{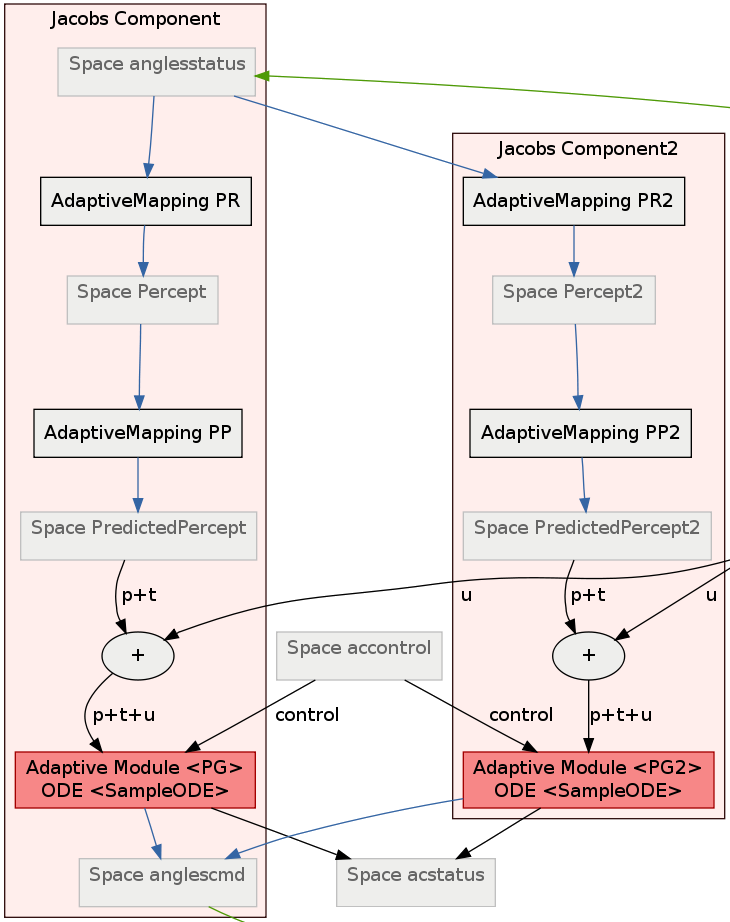}
  \caption{Excerpt of system visualizations of the four motor control blueprints
  introduced in Figure~\ref{fig:example:diff}, expressed in the \amdsl{},
  making different solutions identifiable and comparable.}
 \label{fig:example:dsl}
\end{figure}

Figure~\ref{fig:example:dsl} shows excerpts of system illustrations of all four
approaches, as they were expressed in the \amdsl{} in terms of \texttt{Adaptive
Components}, \texttt{Adaptive Modules} and \texttt{Spaces}. The visualization
was created through the \graphdsl{} as described in
Section~\ref{sec:dsl:generation}, generating GraphML code for rendering.

\section{\uppercase{Conclusion}}
\label{sec:conclusion}
We presented the work in progress of using \dsl{}s for motor control
architectures based on movement primitives in the \amarsi{} project. We
presented, how different notations of architectural blueprints, especially in
inter-disciplinary projects, can prevent comparability of different approaches.
The \amdsl{} expresses motor skill architectures
based on a domain analysis and the resulting architectural
meta-model. As an exemplary use-case we showed how it already helped formulating
different blueprints in the same domain terms and notations.
Similarities of the different approaches could be extracted easier allowing
comparison of the blueprints.

We expect this to be an enabler for architectural research, which was already
found plausible in the first applications as presented in
Section~\ref{sec:application}.
The presented tool-chain based on a state-of-the-art language workbench will enable us to quickly progress
from this starting point towards the core domain aspects of the \amarsi{}
project, namely dynamical systems, learning and adaptive control, eventually
facilitating reproducible research on the architectural level.

Further work will continue to establish a model-driven development process in
\amarsi{} around the presented DSLs, extending the C++ code generation to
verify our approaches on compliant robot platforms.
In terms of further language modularization we will extract the type-system of
the \amdsl{} into a separate \emph{robot data-types} DSL, to
also be accessible for applications outside the motor control
architecture domain, like the meta-model for data interoperability as presented
in~\cite{Wienke2012}.
On a conceptual level further development will continue incorporating dynamic
aspects of motor skills architectures in the \amdsl{}.

\section{\uppercase{Acknowledgement}}
This work was supported by the European Community's Seventh Framework 
Programme FP7/2007-2013 Challenge 2 Cognitive Systems, Interaction, Robotics
 under grant agreement No 248311 - AMARSi.
 
\bibliographystyle{unsrt}
\bibliography{library}

\end{document}